\documentclass{article}

\usepackage[preprint]{neurips_2024}

\usepackage[utf8]{inputenc} 
\usepackage[T1]{fontenc}    
\usepackage{hyperref}       
\usepackage{url}            
\usepackage{booktabs}       
\usepackage{amsfonts}       
\usepackage{nicefrac}       
\usepackage{microtype}      
\usepackage{subcaption}
\usepackage{graphicx}
\usepackage{amsmath}
\usepackage{amsthm}
\usepackage{algorithm}
\usepackage{multirow}
\usepackage[table]{xcolor}
\usepackage{algpseudocode}
\theoremstyle{plain}
\newtheorem{theorem}{Theorem}[section]
\newtheorem{Lemma}{Lemma}[section]

\definecolor{nipsorange}{HTML}{F7B882}
\definecolor{lightnipsorange}{HTML}{FFE4CC}

\hypersetup{
  colorlinks=true,
  linkcolor=black,        
  citecolor=nipsorange,   
  urlcolor=blue
}

\title{Ratio-Variance Regularized Policy Optimization \\for Efficient LLM Fine-tuning}

\author{
  Yu Luo$^1$, \ Shuo Han$^1$, \ Yihan Hu$^1$, \ Dong Li$^{1,}$\thanks{Corresponding author.}, \ Jianye Hao$^2$ \\
  $^1$Department of Foundation Model, 2012 Labs, Huawei\\
  $^2$College of Intelligence and Computing, Tianjin University\\
  \texttt{\{luoyu81, hanshuo15, huyihan6, lidong106\}@huawei.com},\\
  \texttt{jianye.hao@tju.edu.cn}
}

\begin{document}

\maketitle

\begin{abstract}
On-policy reinforcement learning (RL), particularly Proximal Policy Optimization (PPO) and Group Relative Policy Optimization (GRPO), have emerged as the dominant algorithms for fine-tuning Large Language Models (LLMs). While ensuring stability via policy ratio clipping, this heuristic hard constraint imposes a severe cost: it indiscriminately truncates the gradients from high-return yet high-divergence actions—effectively suppressing the model's ``eureka moments'' in complex reasoning. Furthermore, this rigid boundary renders historical data useless once it becomes slightly stale, severely hindering sample efficiency. To overcome these limitations, we propose a more principled approach by revisiting the trust-region objective in policy optimization. We theoretically demonstrate that explicitly constraining the \textit{variance (second central moment) of the policy ratio} serves as a principled and smoother relaxation of hard clipping, effectively stabilizing updates while preserving the learning signal from valuable trajectories. Building on this insight, we introduce \textbf{R$^2$VPO} (Ratio-Variance Regularized Policy Optimization), a novel framework that incorporates this variance constraint using a primal–dual formulation. This design naturally enables R$^2$VPO to achieve stable on-policy performance and, more promisingly, unlocks the potential of off-policy learning in LLMs: by dynamically weighing stale data instead of discarding it, R$^2$VPO enables robust data reuse from a replay buffer. We extensively validate R$^2$VPO on fine-tuning state-of-the-art LLMs—including DeepSeek-Distill-Qwen-1.5B and the openPangu-Embedded series (1B/7B)—across several challenging math reasoning benchmarks. Our results show that R$^2$VPO consistently achieves superior asymptotic performance—with average relative gains of up to \textbf{17\%} over strong clipping-based baselines—while requiring approximately \textbf{50\%} fewer rollouts to reach convergence. These findings establish ratio-variance control as a highly promising direction for simultaneously advancing the stability and data efficiency of RL for LLM alignment.
\end{abstract}

\begin{figure}[htpb]
  \centering
  \includegraphics[width=\textwidth]{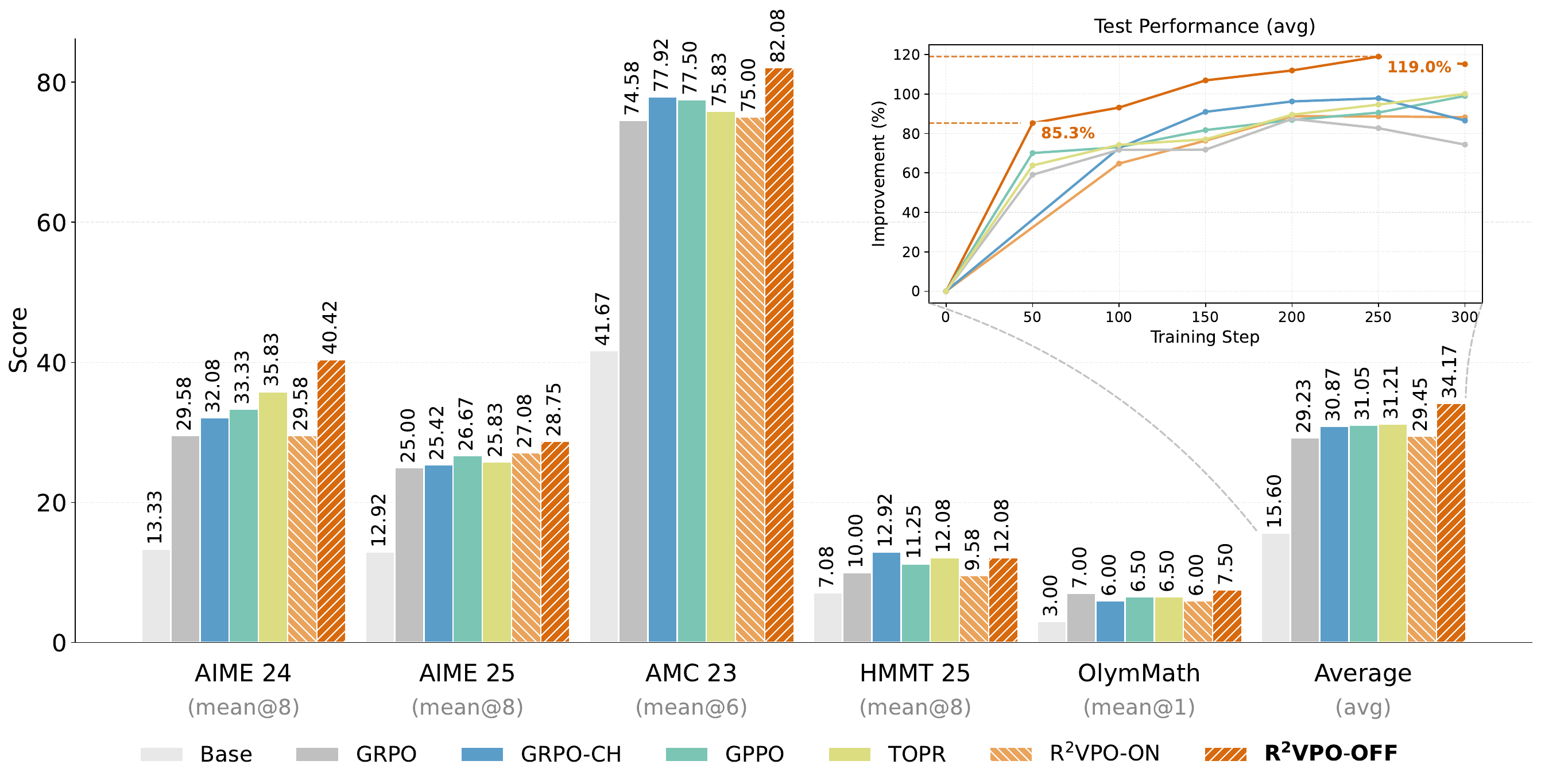}
  \vspace{-10pt}
  \caption{\textbf{R$^2$VPO achieves a \textbf{17\%} average performance gain while requiring \textbf{50\%} fewer data rollouts} compared to strong baselines on DeepSeek-Distill-Qwen-1.5B. The main bar chart details the \textbf{Pass@1} accuracy across five math benchmarks, where the off-policy variant consistently demonstrates superior asymptotic performance. The inset visualizes the training dynamics, highlighting a total improvement of \textbf{119\%} over the base model and rapid convergence that matches the baselines' peak performance in significantly fewer steps.}
  \vspace{-10pt}
  \label{fig:ab}
\end{figure}

\section{Introduction}
Large Language Models (LLMs) have evolved into the cornerstone of modern artificial intelligence~\citep{zhao2023survey,chang2024survey,tang2025pangu}, exhibiting sophisticated capabilities in complex reasoning~\citep{zhou2022least}, instruction following~\citep{zhou2023instruction}, and creative generation~\citep{franceschelli2025creativity}. While pre-training on vast corpora instills general knowledge, Reinforcement Learning (RL) plays a pivotal role in the post-training alignment phase~\citep{ouyang2022training}. By optimizing models against a reward signal—derived either from human feedback or verifiable outcomes—RL incentivizes the generation of higher-quality, more coherent, and logically sound responses, effectively bridging the gap between raw token probability prediction and desirable human-centric behaviors~\citep{ouyang2022training}.

Currently, on-policy policy gradient methods serve as the dominant paradigm for this fine-tuning process, with Proximal Policy Optimization (PPO)~\citep{schulman2017proximal} and Group Relative Policy Optimization (GRPO)~\citep{shao2024deepseekmath} being the de facto standards. The core mechanism of these algorithms relies on the policy ratio, defined as $r_t(\theta) = \frac{\pi_\theta(a_t|s_t)}{\pi_{\text{old}}(a_t|s_t)}$, which measures the divergence between the current policy being optimized and the behavior policy that collected the data. This ratio acts as an importance sampling weight, allowing the algorithm to estimate the gradient of the expected return using sampled trajectories. Managing the magnitude of this ratio is crucial: if the new policy deviates too far from the old one (i.e., the ratio becomes excessively large or small), the variance (and bias) of the gradient estimator increases dramatically, often resulting in unstable updates or catastrophic policy collapse~\citep{schulman2015trust,schulman2017proximal}.

To maintain stability, state-of-the-art methods ubiquitously employ a hard clipping mechanism. This heuristic forcibly truncates the policy ratio within a narrow interval (e.g., $[1-\epsilon, 1+\epsilon]$) to enforce a trust region. While this prevents destructive updates, we argue that hard clipping imposes a rigid ceiling that fundamentally limits learning in two ways. First, it overlooks rare but valuable actions. When the model discovers a ``eureka'' moment—a high-value action driven by initially low-probability tokens—the policy ratio naturally spikes. Hard clipping indiscriminately truncates this signal, effectively zeroing out the gradient for the most informative samples and suppressing the discovery of sophisticated reasoning paths. Second, it severely restricts data efficiency. As the policy evolves, data collected from just a few iterations ago becomes ``stale'', causing its policy ratio to drift outside the clipping range. Consequently, this data is discarded, preventing the use of replay buffers and forcing the algorithm to rely exclusively on expensive, freshly generated samples. While methods like DAPO~\citep{yu2025dapo}, GPPO~\citep{su2025klear} or TOPR~\citep{roux2025tapered} have attempted to relax clipping through asymmetric or gradient-based modifications, they remain heuristics built around the clipping mechanism, rather than replacing it with an explicit distributional constraint.

In this work, we propose a more principled approach to stability that transcends the limitations of hard clipping. We revisit the trust-region constraint in policy optimization~\citep{schulman2015trust} and demonstrate that it can be locally approximated by constraining \textit{the variance of the policy ratio}. Unlike hard clipping enforces a pointwise constraint, variance regularization operates at the distributional level. It naturally scales down updates based on uncertainty without discarding the gradient direction of high-value outliers, and it automatically down-weights stale off-policy data based on divergence rather than rejecting it entirely. Building on this insight, we develop Ratio-Variance Regularized Policy Optimization (R$^2$VPO). By formulating the variance constraint as a Lagrangian dual problem, R$^2$VPO utilizes a primal–dual optimization scheme to jointly maximize rewards and minimize ratio variance, enabling both stable on-policy training and robust off-policy data reuse.

We empirically evaluate R$^2$VPO on a suite of challenging math reasoning benchmarks, including AIME 2024, AIME 2025~\citep{maaaime}, AMC 2023~\citep{aimoamc}, OlymMath~\citep{sun2025challenging}, and HMMT 2025~\citep{harvardmmt}. Using both fast-thinking (openPangu-Embedded-1B) and slow-thinking (DeepSeek-Distill-Qwen-1.5B, openPangu-Embedded-7B) models, we demonstrate that R$^2$VPO consistently outperforms hard-clipping policy-gradient baselines. Notably, our results highlight a distinct advantage in efficiency: the off-policy variant of R$^2$VPO matches the peak performance of baselines using approximately \textbf{50\%} fewer data rollouts, while the on-policy variant demonstrates exceptional stability, avoiding late-stage degradation even on larger-scale 7B models (Figure~\ref{fig:train_dynamics}). We believe R$^2$VPO provides a principled and flexible alternative for LLM post-training—offering both rapid convergence and scalable stability—and holds strong potential for extension to broader decision-making domains.
\section{Preliminary}\label{sec:preliminary}
\paragraph{RL Formulation for LLMs.}
We formulate the post-training refinement of LLMs as a discrete Markov Decision Process (MDP), defined by the tuple $\mathcal{M}=\langle \mathcal{S}, \mathcal{A}, \mathcal{P}, r, \gamma\rangle$. The state space $\mathcal{S}$ consists of the input prompt $q$ and the history of generated tokens. The action space $\mathcal{A}$ corresponds to the model's vocabulary $\mathcal{V}$, where an action $a_t \in \mathcal{A}$ represents the token selected at time step $t$. The policy $\pi_\theta(a_t|s_t)$, parameterized by $\theta$, maps the current state to a probability distribution over $\mathcal{V}$. Given the conditionally deterministic transition dynamics $\mathcal{P}$ of autoregressive generation (conditioned on the sampled action), the environment yields a sparse scalar reward $r(\tau)$ only upon the completion of a trajectory $\tau = (s_0, a_0, \dots, s_T)$. Consistent with the previous alignment practices~\citep{Polaris2025}, we set the discount factor $\gamma=1$ and omit the explicit KL-divergence penalty regarding the reference policy in the reward formulation, focusing purely on maximizing the external task reward. The optimization objective is thus to maximize the expected return:
\begin{equation} \label{eq:objective}
\mathcal{J}(\theta) = \mathbb{E}_{\tau \sim \pi_{\theta}}\left[ r(\tau) \right] = \mathbb{E}_{q \sim \mathcal{D}, o \sim \pi_{\theta}(\cdot|q)}\left[ r(q, o) \right],
\end{equation}
where $\mathcal{D}$ represents the dataset of prompts, and $o$ denotes the generated response.

\paragraph{Group Relative Policy Optimization.}
Group Relative Policy Optimization (GRPO)~\citep{shao2024deepseekmath, guo2025deepseek} has established itself as one of the predominant algorithm for LLMs' post-training. Distinct from standard PPO~\citep{schulman2017proximal}, GRPO eliminates the need for a parametric value function (Critic) to estimate advantages. Instead, for each prompt $q$, it samples a group of $G$ outputs $\{o_1, o_2, \dots, o_G\}$ from the current policy and computes the advantage via group-based normalization. The surrogate objective is defined as
\begin{equation}
\mathcal{J}^{\text{GRPO}}(\theta) = \mathbb{E}_{q \sim \mathcal{D}, \{o_i\}_{i=1}^G \sim \pi_{\text{old}}(\cdot|q)} \left[ \frac{1}{G} \sum_{i=1}^G \sum_{t=1}^{T_i} \min\left( \rho_{i,t}(\theta)\hat{A}_{i,t}, \text{clip}\left(\rho_{i,t}(\theta), 1-\epsilon, 1+\epsilon\right)\hat{A}_{i,t} \right) \right],
\end{equation}
where $\rho_{i,t}(\theta) = \frac{\pi_\theta(a_{i,t}|s_{i,t})}{\pi_{\text{old}}(a_{i,t}|s_{i,t})}$ is the importance sampling ratio. The advantage $\hat{A}_{i,t}$ is derived by normalizing the trajectory rewards within the group:
\begin{equation}
\hat{A}_{i,t} = \frac{r_i - \text{mean}(\{r_1, \dots, r_G\})}{\text{std}(\{r_1, \dots, r_G\}) + \delta},
\end{equation}
where $\delta$ is a small constant for numerical stability. Note that this trajectory-level advantage is broadcast to all tokens in the sequence, following standard practice in GRPO.

While effective in preventing catastrophic updates, the clipping mechanism in GRPO is inherently heuristic. Mathematically, when the policy ratio (importance sampling ratio) $\rho_{i,t}(\theta)$ falls outside the trust region $[1-\epsilon, 1+\epsilon]$, the objective function becomes locally constant with respect to $\theta$, resulting in a vanishing gradient ($\nabla_\theta \mathcal{J} = 0$). This ``dead zone'' effectively discards gradient information from samples with significant policy divergence, regardless of whether that divergence stems from a highly beneficial exploration. This limitation motivates our search for a continuous, variance-aware regularizer that maintains stability without indiscriminately truncating the learning signal.
\section{Ratio-Variance Regularized Policy Optimization}
Our framework begins by revisiting the fundamental objective of Trust Region Policy Optimization (TRPO)~\citep{schulman2015trust}, which seeks to maximize the expected return subject to a strict constraint on the policy update size. In this work, we aim to solve this constrained optimization problem through a more principled theoretical lens, rather than relying on heuristic truncation like the hard clipping mechanism in PPO and GRPO.

To extend the applicability of this framework to scenarios involving off-policy data reuse (e.g., experience replay buffers or asynchronous updates common in LLM training), we generalize the standard behavior policy from the immediate predecessor $\pi_{\text{old}}$ to a broader off-policy distribution $\pi_{\text{off}}$. Here, $\pi_{\text{off}}$ represents the mixture of historical policies that generated the available trajectories in the dataset $\mathcal{D}$. Consequently, our generalized optimization goal is to maximize the importance-sampled objective subject to a divergence constraint:
\begin{equation}
\label{eq:constrained_obj}
\max_{\theta} \ \mathcal{J}(\theta) = \hat{\mathbb{E}}_{(s_t,a_t)\sim\pi_{\text{off}}}\left[\rho_t(\theta)\hat{A}^{\pi_{\text{off}}}(s_t,a_t)\right] \quad \text{s.t.} \quad \hat{\mathbb{E}}_{s_t\sim\pi_{\text{off}}}\left[D\left(\pi_\theta(\cdot|s_t)||\pi_{\text{off}}(\cdot|s_t)\right)\right] \leq \delta,
\end{equation}
where $\rho_t(\theta)=\frac{\pi_\theta(a_t|s_t)}{\pi_{\text{off}}(a_t|s_t)}$ is the generalized importance sampling ratio, $\hat{A}_t$ is the advantage estimate associated with the sample (e.g., the group-relative advantage in GRPO), and $D(\cdot||\cdot)$ is a divergence metric defining the trust region.

The choice of the divergence metric $D$ is critical. Traditional methods typically employ the Forward KL divergence $D_{\text{KL}}(\pi_{\text{off}}||\pi_{\theta})$ or Reverse KL divergence $D_{\text{KL}}(\pi_{\theta}||\pi_{\text{off}})$. Recent studies~\citep{li2025choice,zhang2025design} argue that the asymmetry of KL divergence leads to distinct optimization biases: Forward KL is ``mode-covering'' (potentially conservative), while Reverse KL is ``mode-seeking'' (potentially unstable). To mitigate these extremes, we propose using the Jensen–Shannon (JS) Divergence $D_{\text{JS}}\left(\pi_\theta||\pi_{\text{off}}\right)$, a symmetric and bounded metric defined as:
\begin{equation}
D_{\text{JS}}\left(\pi_\theta||\pi_{\text{off}}\right)=\frac{1}{2}D_{\text{KL}}\left(\pi_\theta\Big\Vert\frac{\pi_\theta+\pi_{\text{off}}}{2}\right)+\frac{1}{2}D_{\text{KL}}\left(\pi_{\text{off}}\Big\Vert\frac{\pi_\theta+\pi_{\text{off}}}{2}\right).
\end{equation}
The JS divergence provides a balanced measure of policy distance. Furthermore, as we will demonstrate in the subsequent derivation, under a second-order Taylor expansion, these divergence metrics (Forward KL, Reverse KL, and JS) become asymptotically equivalent to the variance of the policy ratio, differing only by a constant scaling factor. This insight shows that, in the local trust-region regime where $\pi_\theta$ remains close to $\pi_{\text{off}}$, these divergence metrics induce equivalent second-order constraints on the policy ratio variance, differing only by a constant scaling factor. The resulting symmetric trust-region problem is
\begin{equation} \label{eq:js_problem}
\max_{\theta} \ \mathcal{J}(\theta) \quad \text{s.t.} \quad \hat{\mathbb{E}}_{s_t\sim\pi_{\text{off}}}\left[D_{\text{JS}}\left(\pi_\theta(\cdot|s_t)||\pi_{\text{off}}(\cdot|s_t)\right)\right]\leq\delta.
\end{equation}
This formulation naturally supports off-policy learning by constraining the new policy to stay within a stable radius of the data-generating distribution $\pi_{\text{off}}$, regardless of how ``stale'' that data might be.

\subsection{From JS Divergence to Policy Ratio Variance}
Efficiently enforcing the trust region constraint is a central challenge in RL. Previous works like TRPO~\citep{schulman2015trust} utilize a second-order Taylor approximation of the KL divergence, leading to the computationally expensive Fisher Information Matrix. Other approaches approximate the divergence via higher-order polynomial expansions of the likelihood ratio (e.g., K2/K3 estimators)~\citep{schulman2017proximal}. In this work, we seek a more direct and computationally lightweight statistic. By focusing on the properties of the policy ratio $\rho_t(\theta)$ within the symmetric JS divergence, we derive a closed-form relationship that connects the abstract divergence constraint directly to the variance of the ratio.

\begin{Lemma}[Approximation of JS Divergence]\label{le:js_app}
Let $\rho_t(\theta) = \pi_\theta(a_t|s_t) / \pi_{\text{off}}(a_t|s_t)$ be the policy ratio. For a policy $\pi_\theta$ sufficiently close to $\pi_{\text{off}}$, the expected Jensen-Shannon divergence can be approximated by the second-order moment (variance) of the policy ratio:
\begin{equation}\label{eq:js_constrained}
\hat{\mathbb{E}}_{s_t\sim\pi_{\text{off}}}\left[D_{\text{JS}}(\pi_\theta||\pi_{\text{off}})\right] \approx \frac{1}{8}\hat{\mathbb{E}}_{(s_t,a_t)\sim\pi_{\text{off}}}\left[\left(\rho_t(\theta)-1\right)^2\right] = \frac{1}{8}\text{Var}_{(s_t,a_t)\sim\pi_{\text{off}}}\left[\rho_t(\theta)\right].
\end{equation}
\end{Lemma}

\textit{Proof Sketch.}
The proof relies on the second-order Taylor expansion of the $f$-divergence. The JS divergence is an $f$-divergence with generator function $f(u) = \frac{u}{2}\log u + \frac{1}{2}\log(\frac{2}{u+1})$. We expand $f(\rho_t)$ around $\rho_t=1$ (where $\pi_\theta \approx \pi_{\text{off}}$). The zeroth and first-order terms vanish since $\hat{\mathbb{E}}[\rho_t(\theta)]=1$. The second derivative at equilibrium is $f''(1) = \frac{1}{4}$. Consequently, the divergence is dominated by the quadratic term: $\mathbb{E}[f(\rho_t)] \approx \frac{f''(1)}{2}\mathbb{E}[(\rho_t-1)^2] = \frac{1}{8}\text{Var}[\rho_t]$. This confirms that controlling the variance of the ratio is locally equivalent to enforcing the JS constraint.

It is crucial to note that this quadratic form is structurally invariant across the divergence family. Both Forward and Reverse KL divergences converge to this same variance-centric form under second-order expansion, differing only by constant second-order coefficients, as commonly derived in local approximations of $f$-divergences:
\begin{equation}
\hat{\mathbb{E}}\left[D_{\text{KL}}(\pi_\theta || \pi_{\text{off}})\right] \approx \hat{\mathbb{E}}\left[D_{\text{KL}}(\pi_{\text{off}} || \pi_\theta)\right] \approx \frac{1}{2}\text{Var}\left[\rho_t(\theta)\right].
\end{equation}
Thus, constraining the ratio variance effectively enforces a trust region under any of these divergence metrics, justifying our focus on this tractable statistic.

\begin{figure}[t]
    \centering
    \begin{subfigure}[b]{0.32\textwidth}
        \centering
        \includegraphics[width=\textwidth]{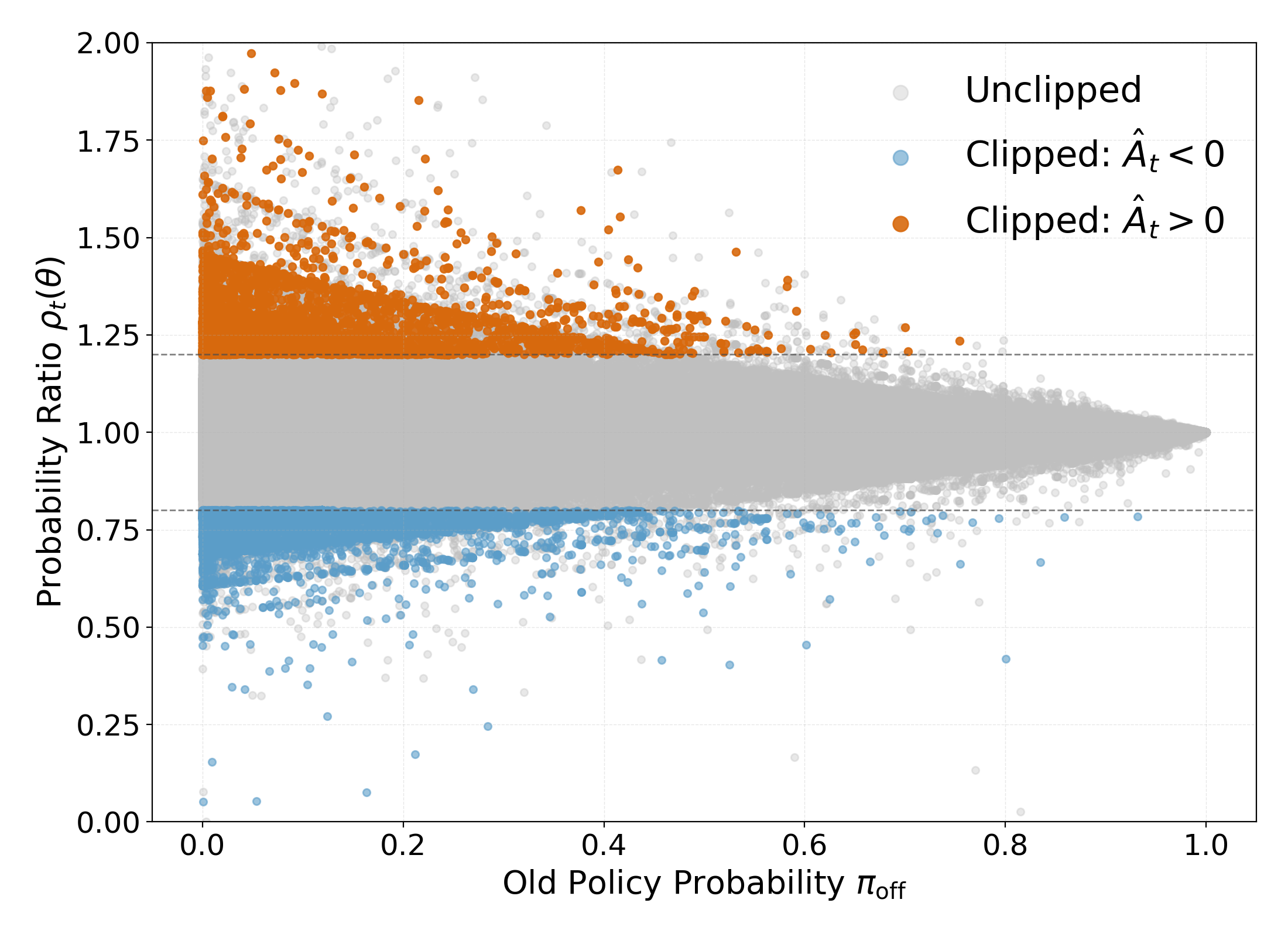}
        \caption{The token distribution}
        \label{fig:clip_range}
    \end{subfigure}
    \hfill
    \begin{subfigure}[b]{0.32\textwidth}
        \centering
        \includegraphics[width=\textwidth]{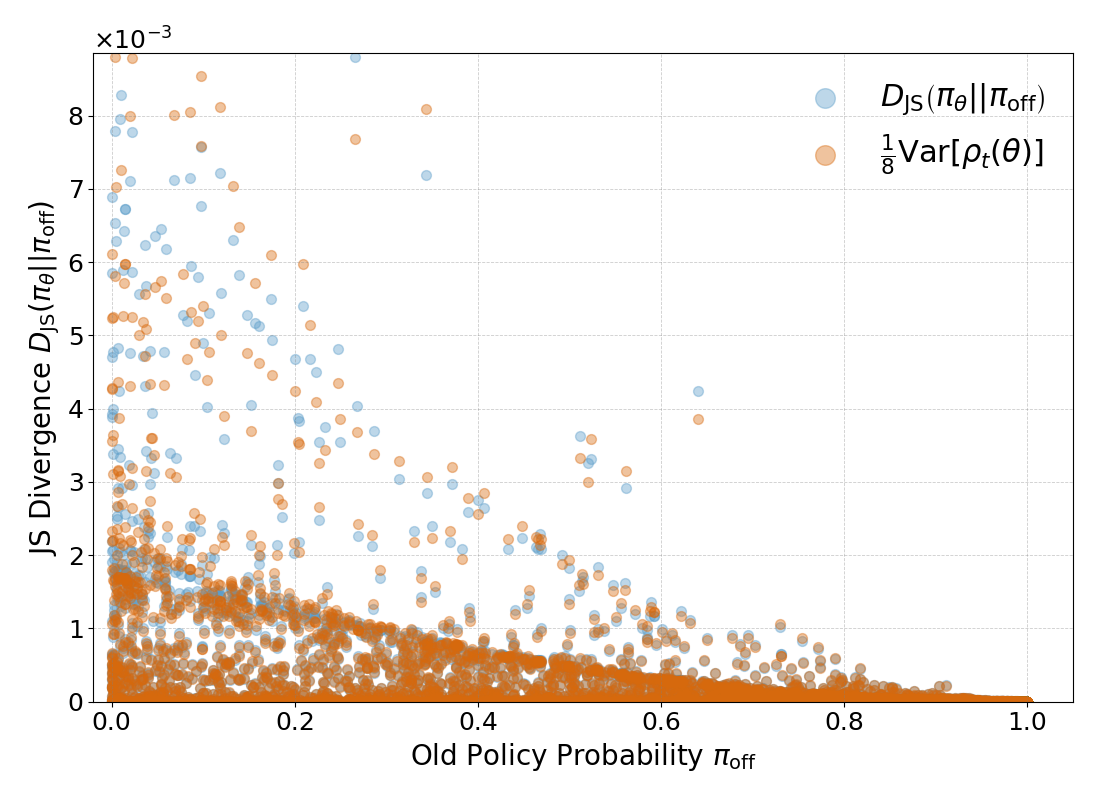}
        \caption{Approximation of JS (by $\pi_{\text{old}}$)}
        \label{fig:JS_old_prob}
    \end{subfigure}
    \hfill
    \begin{subfigure}[b]{0.32\textwidth}
        \centering
        \includegraphics[width=\textwidth]{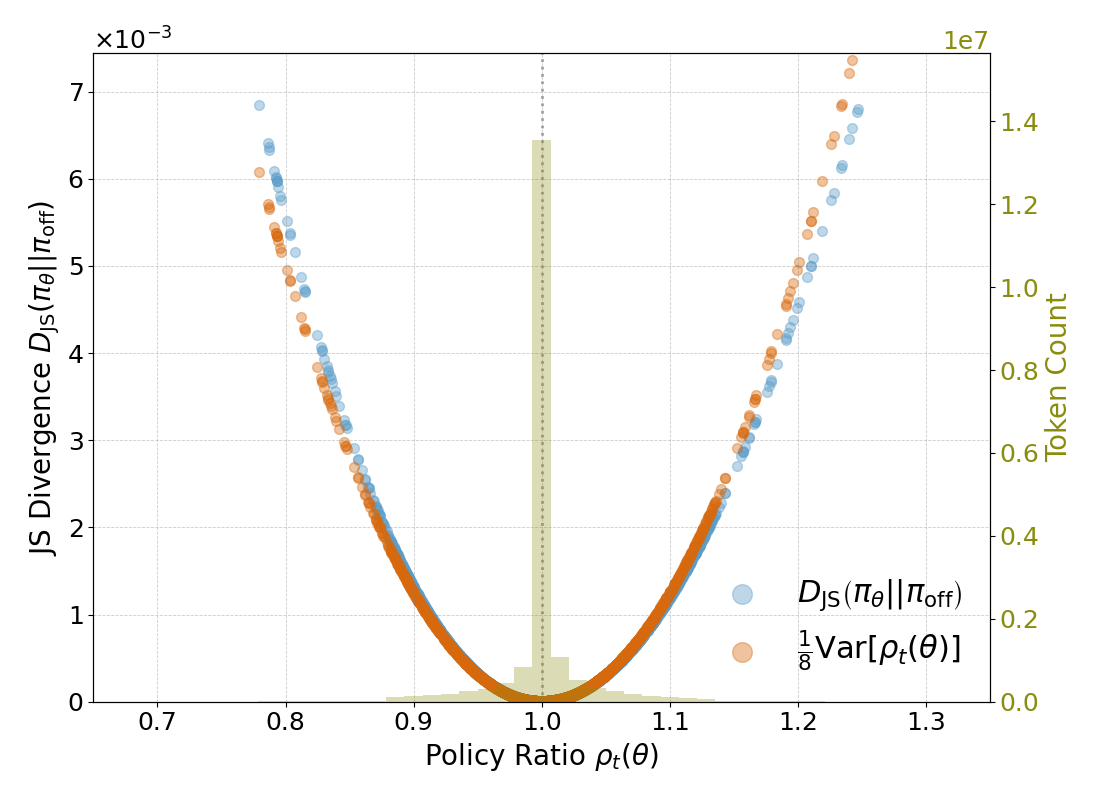}
        \caption{Approximation of JS (by $\rho_\theta(t)$)}
        \label{fig:JS_ratio_variance}
    \end{subfigure}
    \caption{\textbf{Visualization of Policy Ratio Dynamics. Hard clipping indiscriminately truncates high-value exploration, while variance regularization ensures robust stability.} (a) \textbf{The Cost of Clipping:} As shown by the colored points ($|\rho_t - 1| > \epsilon$), standard clipping disproportionately affects low-probability tokens even with positive advantage. This effectively suppresses high-uncertainty yet critical ``eureka moments'' in reasoning. (b) \& (c) \textbf{The Variance Solution:} The proposed variance proxy (orange) demonstrates tight alignment with the true JS divergence (blue). Unlike KL-divergence, it maintains numerical stability even when $\pi_{\text{off}} \to 0$, preventing optimization collapse.}
    \vspace{-10pt}
    \label{fig:three_visualization}
\end{figure}

This variance-based approximation fundamentally shifts the optimization paradigm from heuristic pointwise truncation to principled distributional regularization. While standard hard-clipping mechanisms act as a binary ``switch''—indiscriminately zeroing out gradients for any sample exceeding a fixed threshold—our approach offers a flexible alternative. Specifically, Eq.~\eqref{eq:js_constrained} imposes a \textit{distributional} constraint. \textit{It does not cap individual ratios but regulates the dispersion of the entire batch. A large ratio $\rho_t$ is permitted but incurs a quadratic penalty proportional to its deviation.} This ensures that gradients from high-value outliers are scaled rather than truncated, preserving the direction of improvement while maintaining overall policy stability.

To investigate the impact of hard clipping, we visualize the relationship between token-level policy ratios and the behavior policy's confidence in Figure~\ref{fig:three_visualization}(a), referring to~\citet{wang2025aspo}. The plot reveals a clear structural dichotomy: For tokens where the behavior policy is certain ($\pi_{\text{off}} \approx 1$), the policy ratio $\rho_t$ tightly clusters around $1.0$ (grey region). These tokens represent established knowledge and are rarely affected by clipping. Conversely, tokens with low prior probability ($\pi_{\text{off}} \to 0$) exhibit high ratio variance and dispersion. These ``high-entropy'' tokens often correspond to novel exploration or critical turning points in reasoning paths. However, as shown by the dense colored regions in Figure~\ref{fig:three_visualization}(a), these are precisely the tokens most susceptible to hard clipping. By truncating these ratios, standard methods effectively suppress the learning signal from the most informative parts of the trajectory, hindering the model's ability to correct errors or reinforce new discoveries. Furthermore, Figures~\ref{fig:three_visualization}(b) and (c) empirically validate the fidelity of our variance-based regularizer. The quadratic approximation $\frac{1}{8}(\rho_t - 1)^2$ closely tracks the true Jensen-Shannon divergence, particularly in the high-density regions where the majority of training samples reside. A critical advantage observed here is numerical stability: unlike the Reverse-KL divergence, which can explode when $\pi_{\text{off}} \to 0$ (if the new policy maintains probability), our variance proxy and the bounded JS divergence remain well-behaved. This confirms that R$^2$VPO provides a robust mechanism to enforce trust regions without the instability of KL-based constraints or the information loss of heuristic clipping.

\subsection{Primal-Dual Formulation}
Leveraging the approximation established in Lemma~\ref{le:js_app}, we can now recast the intractable JS-constrained problem into a tractable variance-constrained optimization. By substituting the variance term $\mathbb{E}[(\rho_t(\theta)-1)^2]$ for the divergence constraint in Eq.~\eqref{eq:js_problem}, the optimization objective becomes:
\begin{equation}
\label{eq:var_problem}
\max_{\theta} \ \hat{\mathbb{E}}_{(s_t,a_t)\sim\pi_{\text{off}}}\left[\rho_t(\theta)\hat{A}_t\right] \quad \text{s.t.} \quad \hat{\mathbb{E}}_{(s_t,a_t)\sim\pi_{\text{off}}}\left[(\rho_t(\theta)-1)^2\right] \leq \delta.
\end{equation}

Solving this constrained problem directly is challenging due to the hard boundary on the variance. To address this, we employ the method of Lagrange multipliers to relax the hard constraint into a penalty term. This transforms the problem into an unconstrained saddle-point optimization problem, formalized as follows:

\begin{theorem}[R$^2$VPO Primal-Dual Objective]\label{thm:obj_rvpo}
The constrained optimization problem in Eq.~\eqref{eq:var_problem} is equivalent to the following unconstrained min-max problem via Lagrangian duality:
\begin{equation}
\label{eq:primal_dual}
\min_{\lambda \ge 0} \max_{\theta} \ \mathcal{L}(\theta, \lambda) = \hat{\mathbb{E}}_{(s_t,a_t)\sim\pi_{\text{off}}} \left[ \rho_t(\theta)\hat{A}_t - \lambda \left( (\rho_t(\theta)-1)^2 - \delta \right) \right],
\end{equation}
where $\lambda$ is the dual variable (Lagrange multiplier) controlling the strength of the variance regularization, and $\delta$ represents the target tolerance level for policy deviation.
\end{theorem}

This formulation allows us to jointly optimize the policy parameters $\theta$ and the regularization coefficient $\lambda$. To understand the mechanism of this objective, we analyze the gradient of the Lagrangian $\mathcal{L}$ with respect to the policy parameters $\theta$. By taking the derivative, we obtain:
\begin{equation}
\label{eq:gradient}
\nabla_\theta \mathcal{L}(\theta,\lambda) = \hat{\mathbb{E}}_{(s_t,a_t)\sim\pi_{\text{off}}} \Bigg[ \underbrace{\left( \hat{A}_t - 2\lambda(\rho_t(\theta)-1) \right)}_{\text{Regularized Advantage}} \rho_t(\theta) \nabla_\theta \log \pi_\theta(a_t|s_t) \Bigg].
\end{equation}
The gradient structure in Eq.~\eqref{eq:gradient} reveals the core mechanism of R$^2$VPO. The term $2\lambda(\rho_t(\theta)-1)$ acts as a dynamic, instance-dependent regularizer that modulates the original advantage $\hat{A}_t$.
Unlike hard clipping, which zeroes out the gradient when the ratio exceeds a threshold ($\nabla \mathcal{L}=0$), this formulation preserves the sign of the policy gradient induced by $\hat{A}_t$ but scales its magnitude. Specifically:
\begin{itemize}
    \item When the policy ratio is close to 1 (i.e., $\pi_\theta \approx \pi_{\text{off}}$), the penalty term vanishes, and the update is driven purely by the task advantage $\hat{A}_t$.
    \item When the policy deviates significantly (i.e., $|\rho_t-1|$ is large), the penalty term grows linearly with the deviation, effectively reducing the magnitude of the ``Regularized Advantage''. This provides a \textit{soft brake} that counteracts excessive updates without discarding the learning signal entirely.
\end{itemize}

\subsection{Practical Implementation}
From an implementation perspective, R$^2$VPO can be viewed as a no-clip variant of PPO/GRPO, where the hard clipping operator is replaced by a smooth quadratic penalty derived from the trust-region dual. The theoretical foundation laid out in Theorem~\ref{thm:obj_rvpo} translates directly into R$^2$VPO, a streamlined algorithm that replaces the heuristic clipping mechanism with a principled variance penalty. By substituting the complex clip function and its associated hyperparameters (e.g., clip range) with a simple quadratic term $(\rho_t(\theta)-1)^2$, R$^2$VPO simplifies the code implementation while grounding the update in the dual optimization objective derived in Eq.~\eqref{eq:primal_dual}. This mechanism allows R$^2$VPO to reuse data multiple times effectively, as summarized in Algorithm~\ref{alg:r2vpo_combined}.

\paragraph{On-Policy Training.}
For standard on-policy training, R$^2$VPO operates similarly to PPO/GRPO but with a modified loss function. The dual variable $\lambda$, which governs the strength of the regularization, can be treated in two ways: effectively fixed as a hyperparameter (e.g., $\lambda=0.04$ in our experiments) or dynamically optimized via dual gradient descent to strictly satisfy the variance constraint $\delta$. In the dynamic setting, we initialize $\lambda=0.04$ and update it at each step with a learning rate $\eta_\lambda=10^{-3}$ based on the constraint violation:
\begin{equation}
\lambda \leftarrow \max\left(0, \lambda - \eta_\lambda \left(\delta - \hat{\mathbb{E}}[(\rho_t-1)^2]\right)\right).
\end{equation}
This ensures the policy remains within the trust region without manual tuning.

\paragraph{Off-Policy Training.}
Crucially, the variance-based formulation naturally extends to off-policy training, unlocking significant gains in sample efficiency. Unlike standard on-policy methods that must discard data immediately after an update, R$^2$VPO leverages a \textbf{Replay Buffer} $\mathcal{D}$ to store and reuse historical reasoning trails. During the inference phase, we store the experience tuple $\tau = \langle q, o, \log\pi_{\text{off}}(o|q), r, \hat{A} \rangle$ into $\mathcal{D}$, where $\pi_{\text{off}}$ is the policy at the time of generation. During the training phase, we sample mini-batches uniformly from $\mathcal{D}$ to update the current policy $\pi_\theta$. The variance penalty term $(\rho_t(\theta)-1)^2$ automatically scales the gradient contribution of these samples: fresh data (where $\rho \approx 1$) contributes fully, while stale data (where $\rho$ diverges) is softly penalized, preventing instability.

\begin{algorithm}[t]
\caption{Training Frameworks for R$^2$VPO}
\label{alg:r2vpo_combined}
\begin{small}
\begin{minipage}[t]{0.49\textwidth}
    \textbf{Algorithm 1} R$^2$VPO (On-Policy)
    \hrule height 0.8pt \vspace{2pt}
    \begin{algorithmic}[1]
        \Require Initial policy $\pi_\theta$, constraint $\delta$
        \State Initialize $\lambda \gets 0.04$, learning rate $\eta_\lambda$
        \Repeat
            \State \textcolor{gray}{\textit{// Data Collection}}
            \For{iteration $k = 1, \dots, K$}
                \State Collect trajectories $\tau \sim \pi_{\theta}$
                \State Compute rewards $r$ and advantages $\hat{A}$
                \State \textcolor{gray}{\textit{// Policy Update}}
                \State Calculate ratio $\rho_t = \frac{\pi_\theta(a_t|s_t)}{\pi_{\theta_{\text{old}}}(a_t|s_t)}$
                \State $\mathcal{L} \gets \hat{\mathbb{E}}[\rho_t \hat{A}_t - \lambda(\rho_t - 1)^2]$
                \State Update $\theta \gets \theta + \alpha \nabla_\theta \mathcal{L}$
                \State \textcolor{gray}{\textit{(Optional) // Dual Update}}
                \State $\lambda \gets \max(0, \lambda - \eta_\lambda (\delta - \hat{\mathbb{E}}[(\rho_t-1)^2]))$
            \EndFor
        \Until{max environment steps}
    \end{algorithmic}
\end{minipage}
\hfill
\begin{minipage}[t]{0.49\textwidth}
    \textbf{Algorithm 2} R$^2$VPO (Off-Policy)
    \hrule height 0.8pt \vspace{2pt}
    \begin{algorithmic}[1]
        \Require Policy $\pi_\theta$, Replay Buffer $\mathcal{D}$
        \State Initialize $\lambda \gets 0.04$
        \Repeat
            \State \textcolor{gray}{\textit{// Inference Phase}}
            \For{each batch of query $x$}
                \State Sample $y \sim \pi_\theta(\cdot|x)$, compute $r, \hat{A}$
                \State Store in $\mathcal{D}$: $(x,y,r,\log \pi_{\text{off}}, \hat{A})$
            \EndFor
            \State \textcolor{gray}{\textit{// Training Phase}}
            \For{each gradient step}
                \State Sample batch $b \sim \mathcal{D}$
                \State $\rho_t \gets \frac{\pi_\theta(a_t|s_t)}{\exp(\log\pi_{\text{off}}(a_t|s_t))}$
                \State Update $\pi_\theta$ via $\nabla_\theta \hat{\mathbb{E}}_b[\rho_t \hat{A}_t - \lambda(\rho_t - 1)^2]$
                \State \textcolor{gray}{\textit{(Optional) Update $\lambda$}}
            \EndFor
        \Until{max environment steps}
    \end{algorithmic}
\end{minipage}
\end{small}
\end{algorithm}
\section{Experiment}
In this section, we conduct a comprehensive evaluation to verify the efficacy of R$^2$VPO. Our experiments are designed to answer two primary research questions:
\begin{itemize}
\item \textbf{RQ1 (Asymptotic Performance):} Does ratio variance regularization provide a more stable and effective optimization landscape than heuristic hard clipping, leading to superior reasoning capabilities?
\item \textbf{RQ2 (Sample Efficiency):} Can R$^2$VPO effectively leverage off-policy data via replay buffers to significantly reduce the computational cost (rollouts) required for convergence?
\end{itemize}

\subsection{Experimental Setup}
\paragraph{Training Datasets.}
All models are trained on the DAPO-Math-17K dataset~\citep{yu2025dapo}, a high-quality corpus designed for mathematical reasoning. The dataset spans a wide range of difficulty levels, from elementary algebra to Olympiad-level problems. To ensure reliable and consistent reward signals, we adopt the \texttt{math\_dapo} evaluation module from VERL~\citep{sheng2024hybridflow}, which performs answer normalization and symbolic equivalence checking.

\begin{table}[t]
    \centering
    \caption{Experimental results on five reasoning benchmarks across different model scales. Both on-policy and off-policy variants of R$^2$VPO demonstrate competitive or superior performance compared to strong baselines. The relative improvement ($\color{red}\uparrow$) is calculated with respect to the Base model.}
    \label{tab:llm_performance}
    \resizebox{\textwidth}{!}{
        \begin{tabular}{l c c c c c c}
            \toprule
            Methods & \textbf{AIME 2024} & \textbf{AIME 2025} & \textbf{AMC 23} & \textbf{HMMT Feb} & \textbf{OlymMath} & \textbf{Avg}\\
            \midrule
            \multicolumn{7}{c}{\textbf{openPangu-Embedded-1B}} \\
            \midrule
            \textit{Base}                & 20.83 & 21.67 & 60.00 & 9.59 & 4.00 & 23.22 \\
            GRPO                         & 27.50 & 24.58 & 65.83 & 11.25 & 6.00 & $27.03_{\color{red}{\uparrow16.4\%}}$ \\
            GRPO-CH         & 28.75 & 25.83 & 67.08 & 11.43 & 7.00 & $28.02_{\color{red}{\uparrow20.7\%}}$ \\
            GPPO                         & 30.41 & 25.42 & 62.92 & \textbf{11.67} & 5.50 & $27.18_{\color{red}{\uparrow17.1\%}}$ \\
            TOPR                         & 26.66 & 25.00 & 63.33 & 10.84 & \textbf{7.50} & $26.67_{\color{red}{\uparrow14.9\%}}$ \\
            R$^2$VPO-ON       & \textbf{31.67} & 24.17 & 65.83 & 11.25 & 6.00 & $27.78_{\color{red}{\uparrow19.7\%}}$ \\
            \textbf{R$^2$VPO-OFF}      & 28.75 & \textbf{28.33} & \textbf{67.50} & 11.25 & 5.00 & $\mathbf{28.17_{\color{red}{\uparrow21.3\%}}}$\\
            \midrule
            \multicolumn{7}{c}{\textbf{openPangu-Embedded-7B}} \\
            \midrule
            \textit{Base}                & 51.67 & 42.92 & 76.25 & 24.17 & 13.00 & 41.60  \\
            GRPO                         & 64.58 & 59.17 & 92.92 & 31.25 & 27.00 & $54.98_{\color{red}{\uparrow32.2\%}}$ \\
            GRPO-CH         & 67.92 & \textbf{61.25} & \textbf{93.75} & 32.08 & \textbf{32.50} & $57.50_{\color{red}{\uparrow38.2\%}}$ \\
            GPPO                         & \textbf{70.42} & 60.83 & 91.88 & 28.33 & 29.00 & $56.09_{\color{red}{\uparrow34.8\%}}$ \\
            TOPR                         & 68.33 & 55.00 & 92.81 & 33.33 & 28.50 & $55.60_{\color{red}{\uparrow33.6\%}}$ \\
            \textbf{R$^2$VPO-ON}       & 67.98 & 59.33 & 92.92 & \textbf{35.00} & \textbf{32.50} & $\mathbf{57.55_{\color{red}{\uparrow38.3\%}}}$ \\
            R$^2$VPO-OFF      & 70.00 & 54.58 & 91.67 & 33.33 & 32.00 & $56.32_{\color{red}{\uparrow35.4\%}}$ \\
            \midrule
            \multicolumn{7}{c}{\textbf{DeepSeek-Distill-Qwen-1.5B}} \\
            \midrule

            \textit{Base} & 13.33  & 12.92 & 41.67 & 7.08 & 3.00 & 15.60  \\
            GRPO & 29.58 & 25.00 & 74.58 & 10.00 & 7.00 & $29.23_{\color{red}{\uparrow87.4\%}}$ \\
            GRPO-CH & 32.08 & 25.42 & 77.92 & 12.92 & 6.00 & $30.87_{\color{red}{\uparrow97.9\%}}$ \\
            GPPO & 33.33 & 26.67 & 77.50 & 11.25 & 6.50 & $31.05_{\color{red}{\uparrow99.4\%}}$ \\
            TOPR & 35.83 & 25.83 & 75.83 & \textbf{12.08} & 6.50 & $31.21_{\color{red}{\uparrow100.1\%}}$ \\
            R$^2$VPO-ON & 29.58 & 27.08 & 75.00 & 9.58 & 6.00 & $29.45_{\color{red}{\uparrow88.8\%}}$ \\
            \textbf{R$^2$VPO-OFF} & \textbf{40.42} & \textbf{28.75} & \textbf{82.08} & \textbf{12.08} & \textbf{7.50} & $\mathbf{34.17_{\color{red}{\uparrow 119\%}}}$ \\
            \bottomrule
        \end{tabular}
    }
\end{table}

\paragraph{Base Models \& Baselines.} We evaluate our method across three LLMs representing different scales and reasoning paradigms: (1) \textbf{DeepSeek-Distill-Qwen-1.5B}~\citep{shao2024deepseekmath}: A slow-thinking model distilled for chain-of-thought reasoning; (2) \textbf{openPangu-Embedded-1B}~\citep{chen2025pangu}: A fast-thinking model optimized for efficient inference; (3) \textbf{openPangu-Embedded-7B}~\citep{chen2025pangu}: A larger-scale slow-thinking model to test scalability.

We compare R$^2$VPO against a suite of state-of-the-art policy gradient methods that employ different strategies to manage the policy ratio: (1) \textbf{GRPO}~\citep{shao2024deepseekmath}: The critic-free variant of PPO, serving as the standard baseline with symmetric hard clipping; (2) \textbf{GRPO-CH (Clip-High)}~\citep{yu2025dapo}: A variant that increases the upper clipping threshold to encourage broader exploration; (3) \textbf{GPPO}~\citep{su2025klear}: A method that preserves gradient direction but not magnitude even when the ratio exceeds clipping bounds; (4) \textbf{TOPR}~\citep{roux2025tapered}: An off-policy efficient method that replaces clipping with asymmetric estimators to handle stale weights.

\paragraph{Benchmarks.}
To assess generalization, we evaluate all methods on six widely used math reasoning benchmarks: AIME 2024, AIME 2025, AMC 2023, HMMT Feb 2025, and OlymMath. These datasets cover diverse domains including algebra, geometry, number theory, and probability. We report the \textbf{Pass@1} accuracy averaged over multiple independent generation runs for each test sample to reduce variance.

\paragraph{Implementation Details.}
All models are trained with a global batch size of $128$ and a group size of $G=8$ responses per query. The learning rate is set to $1\times10^{-6}$, and the maximum sequence length is $16384$ tokens. During inference, we use a temperature of $0.6$ and top-$p$ of $0.95$. Baselines follow the hyperparameter configurations recommended in their original implementations.

For R$^2$VPO, we consider both \textit{on-policy} and \textit{off-policy} training regimes. In the on-policy setting, fresh rollouts are collected at each iteration and discarded after a single update. In the off-policy setting, we employ a FIFO replay buffer with a capacity of four iterations, storing responses, rewards, advantages, and the log probabilities under $\pi_{\text{old}}$. The update-to-data (UTD) ratio is set to $2$, and samples are drawn uniformly from the buffer.

\begin{figure}[t]
    \centering
    \includegraphics[width=\linewidth]{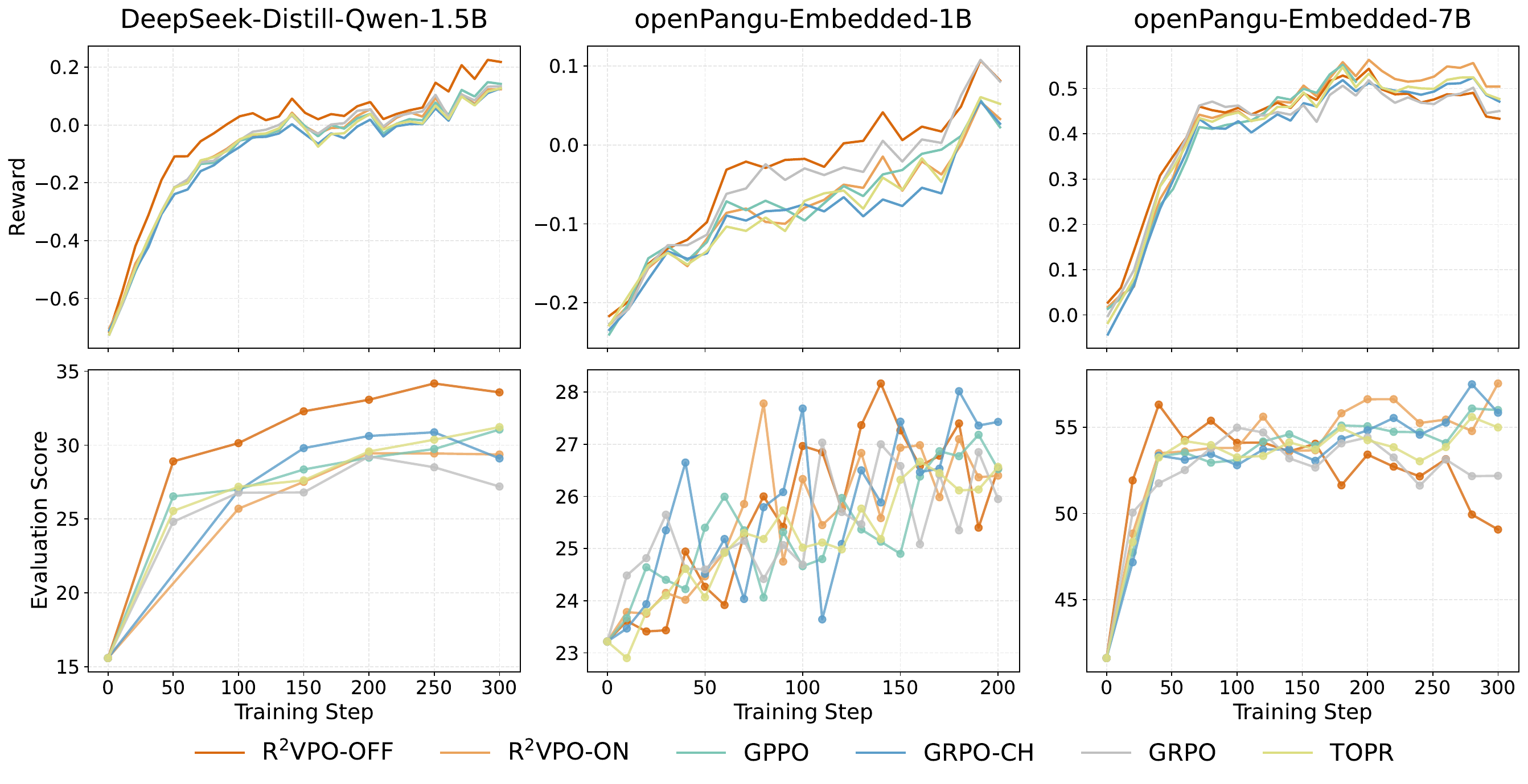}
    \vspace{-15pt}
    \caption{\textbf{Training Dynamics and Evaluation Results.} The learning curves illustrate the evolution of test accuracy and reward throughout the training process. R$^2$VPO (orange lines) consistently achieves faster convergence and higher asymptotic performance compared to baseline methods, particularly in the off-policy setting where data efficiency is maximized.}
    \label{fig:train_dynamics}
\end{figure}

\subsection{Experimental Results}

The results in Table~\ref{tab:llm_performance} show that R$^2$VPO consistently outperforms all strong baselines across different model scales and benchmarks. Replacing heuristic hard clipping with ratio variance regularization provides a more stable and informative optimization landscape, allowing models to better exploit rare high-reward trajectories. For example, on the DeepSeek-Distill-Qwen-1.5B model, R$^2$VPO achieves an improvement of up to \textbf{119\%} over the Base model, and delivers an average gain of \textbf{16.9\%} over the clip-based baselines, with particularly pronounced benefits on challenging benchmarks such as AIME 2024 (improving from 29.58\% to 40.42\%). These results confirm that variance regularization effectively unlocks the models' reasoning capabilities and enhances asymptotic performance (RQ1).

Regarding data efficiency (RQ2), Figure~\ref{fig:train_dynamics} illustrates the training dynamics across different scales. The off-policy variant consistently achieves faster convergence, reaching the peak performance of GRPO with approximately \textbf{50\% fewer rollouts}, while maintaining competitive final scores. On larger models such as openPangu-Embedded-7B, the off-policy approach accelerates early learning but shows mild fluctuations in later stages, whereas the on-policy variant converges more gradually but steadily, ultimately achieving the highest asymptotic performance. This demonstrates that R$^2$VPO provides a flexible trade-off: off-policy training maximizes sample efficiency, while on-policy training ensures stability and consistent improvement at scale.
\section{Related Work}
\paragraph{Trust-Region Alignment and the Evolution of Clipping.}
Post-training LLMs to maximize verifiable rewards has become a critical pathway for enhancing reasoning capabilities~\citep{shao2024deepseekmath, wang2024math}. The theoretical foundation for stable policy improvement lies in Trust Region Policy Optimization (TRPO)~\citep{schulman2015trust}, which enforces a strict KL-divergence constraint to guarantee monotonic improvement. Due to the prohibitive computational cost of second-order optimization, first-order approximations such as PPO~\citep{schulman2017proximal} and its critic-free variant GRPO~\citep{guo2025deepseek} became prevalent. These methods employ a heuristic policy ratio clipping mechanism to approximate the trust region.

While PPO and GRPO have shown strong performance, their ``hard'' clipping truncates gradients once the policy ratio exceeds a fixed threshold, discarding signals from high-value, low-probability tokens (``aha'' moments). Variants like DAPO~\citep{yu2025dapo}, TOPR~\citep{roux2025tapered}, and GPPO~\citep{su2025klear} attempt to mitigate this through relaxed or asymmetric clipping or gradient modifications. More recently, Simple Policy Optimization (SPO)~\citep{xie2024simple} proposes a surrogate objective that tightens PPO’s clipping, improving empirical stability. However, SPO remains a heuristic reformulation of the on-policy objective: it does not provide an explicit trust-region constraint, nor a unifying theoretical perspective across divergence measures, and it treats policy divergence implicitly rather than capturing its distributional geometry. Moreover, SPO has not been empirically validated in large-scale LLM fine-tuning, leaving its practical efficacy in this domain untested. In contrast, R$^2$VPO departs from the clipping paradigm entirely, approximating the trust region via the variance of the policy ratio. This distributional regularization naturally scales gradient magnitudes, preserves all learning signals including rare high-value events, and enables principled off-policy data reuse, as demonstrated by our extensive experiments across multiple LLM architectures and reasoning benchmarks.

\paragraph{Sample Efficiency and Off-Policy Learning.}
In standard RL, off-policy algorithms such as Soft Actor-Critic (SAC)~\citep{haarnoja2018soft} and TD3~\citep{fujimoto2018addressing} achieve superior sample efficiency by aggressively reusing historical experiences stored in replay buffers. Applying off-policy learning to LLM fine-tuning is challenging due to the high dimensionality of language space and distributional staleness. Existing approaches often require heavy Importance Sampling (IS) corrections (e.g., V-trace~\citep{espeholt2018impala}) that introduce high variance, or training a separate Critic Network to estimate Q-values, which doubles memory usage and risks instability. R$^2$VPO circumvents these bottlenecks: it is fully critic-free and enables robust off-policy learning (experimentally validated with staleness of 4) purely through variance-based regularization, achieving the efficiency of off-policy methods without additional architectural complexity.

\section{Conclusion}
We introduced Ratio-Variance Regularized Policy Optimization (R$^2$VPO), a principled framework that addresses the inefficiencies of heuristic hard clipping in LLM post-training. By revisiting the theoretical foundations of trust-region constraints, we showed that explicitly regularizing the \textit{variance} of the policy ratio provides a distributional alternative to pointwise truncation. This approach preserves gradient signals from rare but high-value actions and naturally supports off-policy learning under distributional shifts. Our primal–dual formulation operationalizes these insights into a practical, critic-free algorithm. Empirical results across diverse reasoning benchmarks and model scales—covering both fast- and slow-thinking models—demonstrate that R$^2$VPO consistently outperforms strong hard-clipping baselines. In particular, R$^2$VPO enables effective use of replay buffers, achieving higher asymptotic performance with substantially fewer rollouts. These findings establish ratio-variance control as a promising direction for scalable, data-efficient alignment. Future work will explore extensions to complex decision-making settings, including multi-turn agentic workflows~\citep{zhang2025landscape} and embodied AI~\citep{roy2021machine}.

\bibliography{ref}
\end{document}